\documentclass[10pt,twocolumn,letterpaper]{article}

\pdfoutput=1
\usepackage{adjustbox}	
\usepackage{booktabs}
\usepackage{amsmath,bm}
\usepackage{mathrsfs}
\usepackage{amsthm}
\usepackage{cvpr}
\usepackage{times}
\usepackage{epsfig}
\usepackage{graphicx}
\usepackage{amssymb}
\usepackage{caption}
\usepackage{subcaption}

\usepackage[pagebackref=true,breaklinks=true,letterpaper=true,colorlinks,bookmarks=false]{hyperref}

\cvprfinalcopy

\ifcvprfinal\fi

\begin{document}

\title{Multimodal Visual Concept Learning with Weakly Supervised Techniques}

\author{Giorgos Bouritsas, Petros Koutras, Athanasia Zlatintsi and Petros Maragos \\
School of E.C.E., National Technical University of Athens, Greece\\
{\tt\small gbouritsas@gmail.com, \{pkoutras, nzlat, maragos\}@cs.ntua.gr}
}

\maketitle
\thispagestyle{empty}

\begin{abstract}
Despite the availability of a huge amount of video data accompanied by descriptive texts, it is not always easy to exploit the information contained in natural language in order to automatically recognize video concepts. Towards this goal, in this paper we use textual cues as means of supervision, introducing two weakly supervised techniques that extend the Multiple Instance Learning (MIL) framework: the Fuzzy Sets Multiple Instance Learning (FSMIL) and the Probabilistic Labels Multiple Instance Learning (PLMIL). The former encodes the spatio-temporal imprecision of the linguistic descriptions with Fuzzy Sets, while the latter models different interpretations of each description's semantics with Probabilistic Labels, both formulated through a convex optimization algorithm. In addition, we provide a novel technique to extract weak labels in the presence of complex semantics, that consists of semantic similarity computations. We evaluate our methods on two distinct problems, namely face and action recognition, in the challenging and realistic setting of movies accompanied by their screenplays, contained in the COGNIMUSE database. We show that, on both tasks, our method considerably outperforms a state-of-the-art weakly supervised approach, as well as other baselines.
\end{abstract}

\section{Introduction}

Automatic video understanding has become one of the most essential and demanding challenges and research directions. The problems that span from this field, such as activity recognition, saliency and scene analysis, comprise
detecting events and extracting high level semantics in realistic video sequences. So far, the majority of the methods designed for these tasks deal with visual data ignoring the presence of other modalities, such as text and sound. Nonetheless, the exploitation of the information they provide can lead to better understanding of the underlying semantics. In addition, most of these techniques are fully supervised and are trained on diverse and usually large-scale datasets. Recently, in an attempt to avoid the significant cost of manual annotation, there has been an increasing interest in exploring learning techniques that reduce human intervention.

\begin{figure}[t]
\centering
 \includegraphics[width=\linewidth,scale=0.7]{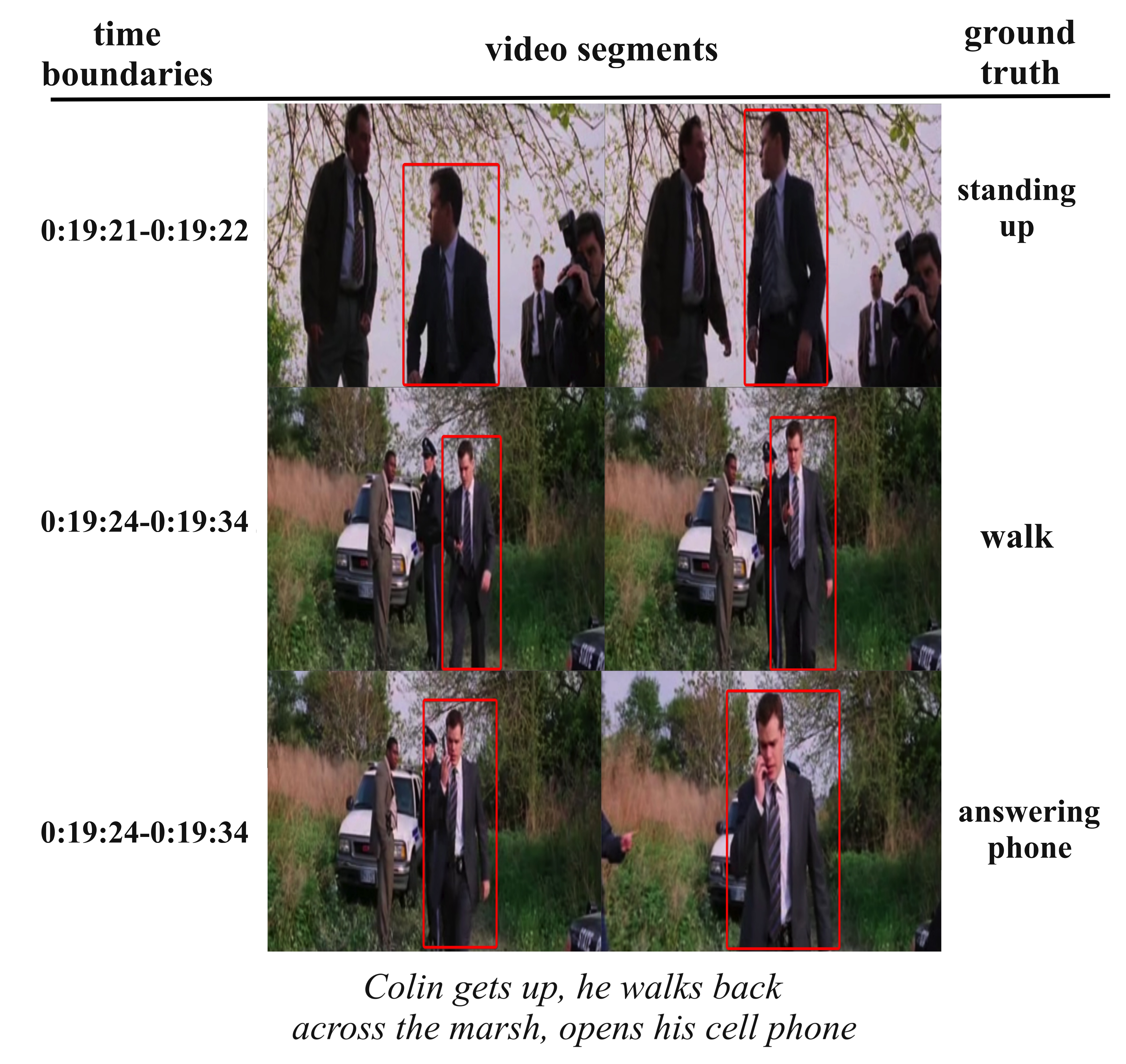}
 \caption{Example of a video segment described by the text shown below the pictures. During the time interval [0:19:21 - 0:19:34] three actions take place (``standing up'', ``walk'', ``answering phone'') performed by the same person (Colin). The corresponding text mentions the actions as ``gets up'', ``walks back'' and ``opens his cell phone'', respectively.}
\vspace{-0.4cm}
\label{fig:fig1}
\end{figure}

Motivated by the above, in this paper we approach video understanding multimodally, where our goal is to recognize visual concepts mining their labels from an accompanying descriptive document. Visual concepts could be loosely defined as spatio-temporally localized video segments that carry a specific structure in the visual domain, which allows them to be classified in various categories.
Some specific examples are human faces, actions, scenes, objects etc. The main reason for using text as a complementary modality is the convenience that natural language provides in expressing semantics. Nowadays, there is a plethora of video data with natural language descriptions, \ie  videos on YouTube \cite{motwani2012improving,naha2016beyond,DBLP:journals/corr/WangKRMSYSCETZZ17}, TV broadcasts including captions \cite{bredin2016multimodal}, videos from parliament or court sessions accompanied by transcripts \cite{maji2007fast} and TV series or movies accompanied by their subtitles, scripts, or audio descriptions \cite{Bojanows13finding,cour2009learning,evangelopoulos2013multimodal,conf/bmvc/EveringhamSZ06,laptev2008learning,rohrbach2016movie,CognimuseDB}. The last category has recently gathered much interest, mainly because of the descriptiveness of these texts and the realistic nature of the visual data. Inspired by such work, we apply our algorithms to movies accompanied by their scripts. In Figure \ref{fig:fig1} we illustrate an example that was extracted from a movie, in which different instances of the action visual concept are described by an accompanying text segment.

Towards this goal, we use a unidirectional model, where information flows from text to video data. This is modeled in terms of weak supervision, while no prior knowledge is used. Specifically,  in order to extract the label from the text for each instance of a visual concept, we face two distinct problems. (i) The first is the absence of specific spatio-temporal correspondence between visual and textual elements. In particular, in the tasks mentioned above, the descriptions are never provided with spatial boundaries and the temporal ones are usually imprecise. (ii) The second major issue is the semantic ambiguity of each textual element. This means that, when it comes to inferring complex semantics from the video such as actions or emotions, the extraction of the label from the text is no longer a straightforward procedure. For example, various expressions could be used to describe the action labeled as ``walking'', such as ``lurching'' or ``going for a stroll''.

Most of the work so far has dealt only to an extent with the spatio-temporal ambiguity, while the semantic one was totally ignored \cite{Bojanows13finding,duchenne2009automatic,laptev2008learning}. In this work, we introduce two novel weakly supervised techniques extending the Multiple Instance Learning (MIL)\footnote{In this paper, the term MIL does not concern only binary classification problems with positive and negative bags, as in its original definition \cite{DIETTERICH199731}, but also the multi-class case.} discriminative clustering framework of \cite{Bojanows13finding}. The first one accounts for the temporal ambiguity variations, which are modeled by Fuzzy Sets (Fuzzy Sets MIL - FSMIL), while the second models the semantic ambiguities by probability mass functions (p.m.f) over the label set (Probabilistic Labels MIL - PLMIL). To the best of our knowledge, this is the first time that both methods are formulated in the context of MIL. In addition, we propose a method of extracting labels in complex tasks using semantic similarity computations. We further improve the recognition, from the perspective of visual representations using features learned from pre-trained deep architectures. The combination of all the above ideas leads to superior performance compared to previous work.

Finally, we focus on the recognition of faces and actions and the evaluation is performed on the COGNIMUSE database \cite{CognimuseDB}. It is important to mention that our methods can be applied to other categories of concepts as long as they can be explicitly described in both modalities {(video \& text)}. 

\section{Related Work}
During the last few years there have been various approaches of understanding videos or images using natural language. Specifically, many have approached the problem as machine translation, such as in \cite{duygulu2002object}, where image regions are matched to words of an accompanying caption and in \cite{naha2016beyond,rohrbach2016movie}, where representations that translate video to sentences and vice-versa are learned. Others have tackled it using video-to-text alignment algorithms \cite{bojanowski:hal-01154523,Sankar09}.

Several works have considered text as means of supervision. In the problem of naming faces, Berg \etal \cite{berg2005s,berg2004names} use Linear Discriminant Analysis (LDA) followed by a modified k-means, to classify faces in newspaper images, while the labels are obtained from captions. In \cite{Bojanows13finding,cour2009learning,cour2011learning,conf/bmvc/EveringhamSZ06,Parkhi15a,ramanathan2014linking,conf/cvpr/SivicEZ09} the authors tackle a similar problem classifying faces in TV series or movies using the names of the speakers provided by the corresponding scripts. The proposed methods are based either on semi-supervised alike techniques using exemplar tracks \cite{conf/bmvc/EveringhamSZ06,conf/cvpr/SivicEZ09}, ambiguous labeling \cite{cour2009learning,cour2011learning} or MIL \cite{Bojanows13finding,Parkhi15a,ramanathan2014linking}. 

The problem of automatically annotating actions in videos has recently drawn the attention of several researchers, because of the need to create diverse and realistic datasets of human activities. For this purpose, Laptev \etal used movie scripts to collect and learn realistic actions \cite{laptev2008learning}. Later on, this work has been improved by incorporating information from the context, leading to the creation of the Hollywood2 dataset \cite{marszalek2009actions}, and by a more accurate temporal localization using MIL \cite{duchenne2009automatic}. In these, a Bag-of-Words text classifier is trained with annotated sentences in order to locate specific actions in the scripts. On the contrary, our work is based only on semantic similarity eliminating the cost of annotation. In Bojanowski \etal \cite{Bojanows13finding}, MIL is also used to jointly learn names and actions, while in Miech \etal \cite{miech2017learning}, the algorithm is improved allowing large-scale optimization via a variant of the Block-Coordinate Frank-Wolfe algorithm. In \cite{miech2017learning}, a supervised approach is once more followed for the extraction of labels from the text, contrary to \cite{Bojanows13finding}, where SEMAFOR \cite{das2012exact} is used, a semantic role labeling parser, searching for two action frames. This unsupervised method, despite its promising results, cannot be easily generalized to custom actions. Similarly, in \cite{shah2016weakly}, the authors propose methods for learning multiple concepts jointly, introducing an extension of the Indian Buffer Process that is constrained by the information provided by the text. All the above end up in considering only the most certain labels that the text provides, ignoring possible paraphrases or synonyms. This allows an automatic collection of data with limited noise, but in general it leads to understanding a small proportion of each individual video.

In order to learn from partially labeled data, there has been an extensive study on weakly supervised classification \cite{hernandez2016weak}. Learning with probabilistic labels has been examined in \cite{jin2003learning} under a probabilistic framework. Cour \etal \cite{cour2009learning} formulated a sub-category of this method, where all possible labels are distributed uniformly (candidate labels) and the classification is performed by minimizing a convex loss function. Both papers concern a single instance setting, namely a p.m.f over the label set is assigned to individual instances. On the contrary, we assign a p.m.f to bags-of instances, generalizing previous formulations. MIL has been largely studied in the machine learning community starting from Diettrich \etal \cite{DIETTERICH199731}, where drug activity was predicted.  Except for the efforts on naming faces mentioned before, MIL has been used in detecting objects \cite{zhang2006multiple} and classifying scenes \cite{maron1998multiple} in images, where annotation lacks specific spatial localization.  While the definition of MIL is sufficient for most of its applications, it is important sometimes to make discriminations between instances in each bag.  In order to model this case, we redefine MIL using Fuzzy Sets.

\section{Multimodal Learning of Concepts}
Given a video and a descriptive text that are crudely aligned \cite{conf/bmvc/EveringhamSZ06}, namely each portion of the text is attributed with temporal boundaries, we aim to localize and identify all the existing instances of a chosen visual concept, such as faces, actions or scenes.  The adversities of such a task are clearly illustrated in Figure \ref{fig:fig1}, where the concept examined is that of human actions. Our approach breaks down the problem into three subproblems. (a) First of all, the exact position in space and time of each visual concept is unknown, thus it needs to be detected automatically.  (b) Secondly, concepts are usually expressed in the text in a different way than their original definition. For instance, as shown in Figure \ref{fig:fig1}, the action ``standing up'' is mentioned by the phrase ``gets up'', while the action ``answering phone'' is mentioned by the phrase ``opens his cell phone''. In order to tackle this problem, we need to detect the part of the text that implies a concept and then mine the label information.  (c) Finally, following the alignment procedure, the text is divided into segments that describe specific time intervals of the video. Each one of them might mention more than one instances of a visual concept. Thus, we need to apply a learning procedure that matches the mined labels with the detected concepts. Note here that sometimes a concept described in the text might not appear in the video or vice-versa. As a result, we need to design an algorithm that learns the visual concepts globally without restricting each one of them to the labels mentioned in its corresponding time interval.

Solving (a) and (b) requires task dependent systems, which are both described in section \ref{section:exp}. The outputs of these systems are perceived as visual and linguistic objects ($v$ and $w$, respectively) with their temporal boundaries determined. 
Following the computation of these, we address (c) and we formulate the learning algorithms.

\subsection{Problem Statement}
\begin{figure}[t]
\includegraphics[width=\linewidth]{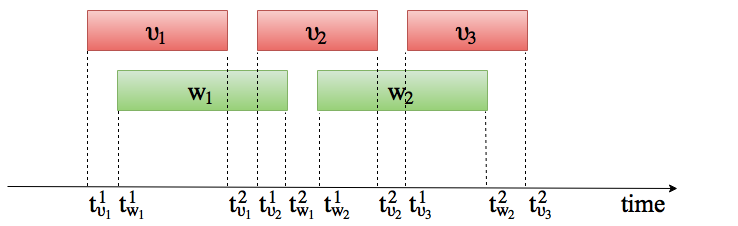}
 \caption{Illustration of the two modalities as parallel data streams.}
\vspace{-0.4cm}   
\label{fig:fig2}
\end{figure}

We assume a dual modality scheme, where both modalities carry the same semantics. This can be modeled with two data streams flowing in parallel as time evolves (Figure \ref{fig:fig2}). The first data stream consists of the unidentified visual objects that we want to recognize. We denote as $\mathcal{V}$ the set of $V$ visual objects. The second modality consists of the linguistic objects that carry in some way the information for the identification of each $v\in\mathcal{V}$, namely they describe the $v$. We denote as $\mathcal{W}$ the set of $W$ linguistic objects (\ie words or sentences). 

\begin{figure*}[h]
    \centering
    \begin{subfigure}[b]{0.45	\textwidth}
        \centering
        \includegraphics[width=1\linewidth]{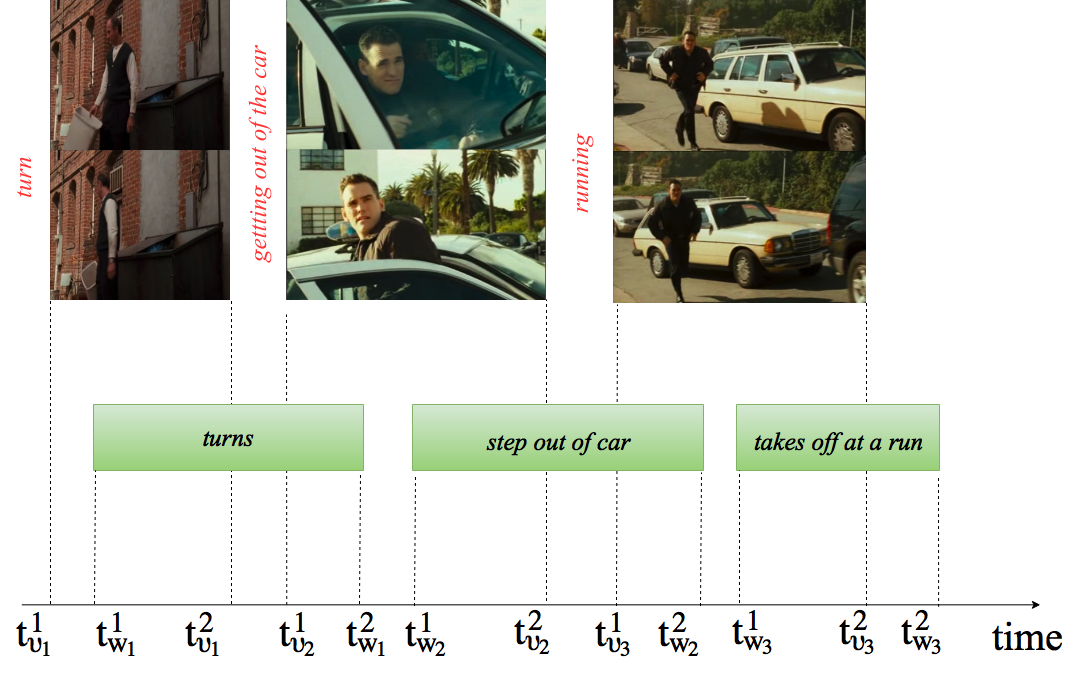}
        \caption{}
    \end{subfigure}
    \begin{subfigure}[b]{0.45\textwidth}
        \centering
        \includegraphics[width=1\linewidth]{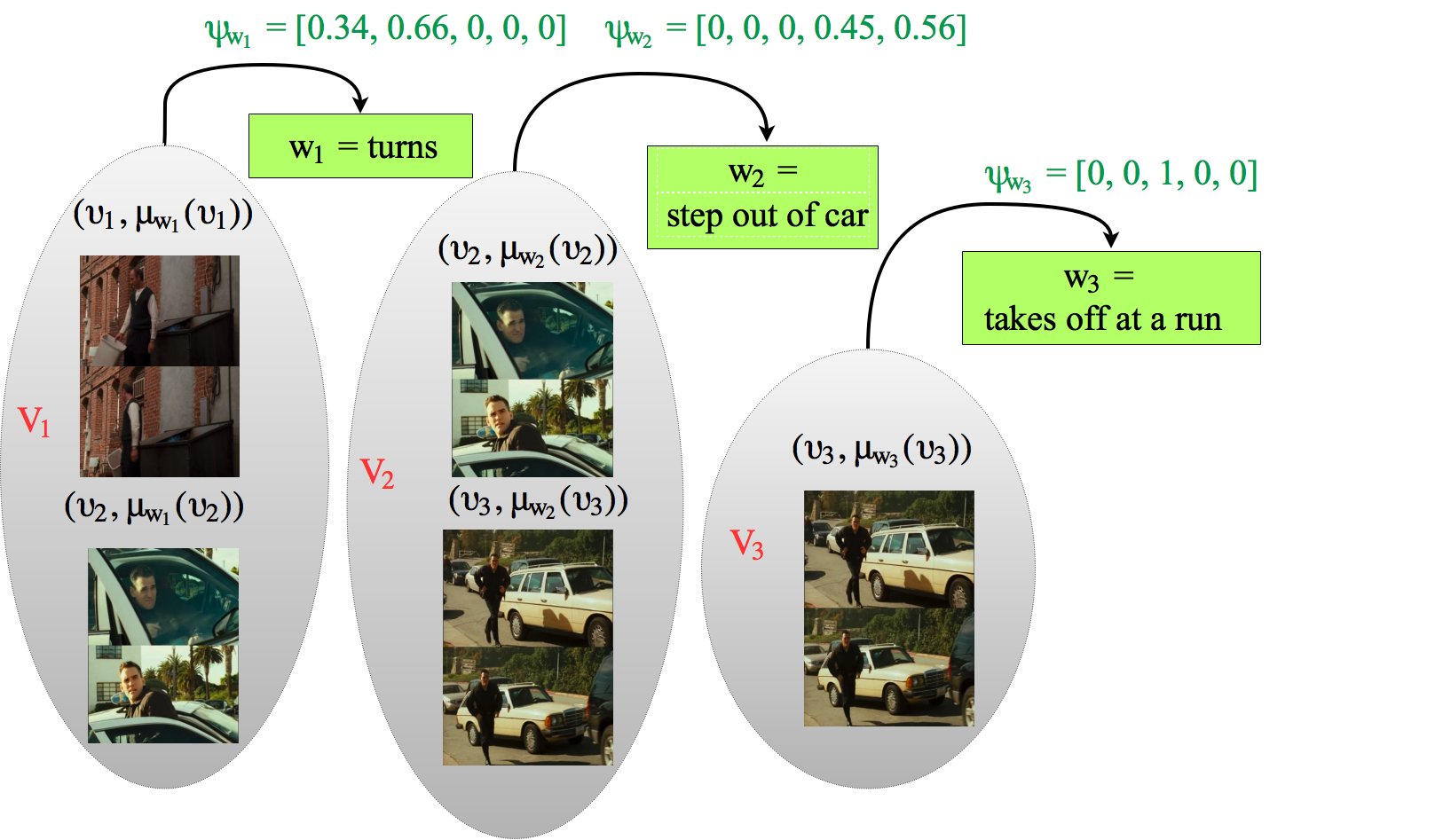}
        \caption{}
    \end{subfigure}
    \vspace{-0.2cm}
    \caption{\textbf{(a)}: Example of the two data streams formed by linguistic and visual objects concerning the concept of human actions. Next to each visual object we demonstrate its ground truth. The formation of the streams is carried out by solving subproblems (a) and (b). \textbf{(b)}: The construction of bags under the MIL setting with Fuzzy Sets and Probabilistic Labels. The label set here is [\textit{``walk'', ``turn'', ``running'', ``driving car'', ``getting out the car''}].}
    \vspace{-0.2cm}
    \label{fig:fig3}

\end{figure*}

We assume that each $v$ is represented in a feature space and its representation is a vector $\bm{x}_v$. 
We define a matrix ${\bm{X}\in \mathbb{R}^{V\times D}}$ containing all the visual features. The time interval of each $v$ is denoted as $T_v=[t_v^1,t_v^2]$.

Let $\mathcal{Y}$ be the label set of $Y$ discrete labels. Each $w\in\mathcal{W}$ is mapped to a label $y_w$ through a mapping $\psi : \mathcal{W} \to \mathcal{Y}$. This can be either deterministic, matching each $w$ to a sole label $y$, or probabilistic, matching each $w$ to a p.m.f over the label set (see section \ref{probabilistic}). The time interval of each $w$ is denoted as $T_w=[t_w^1,t_w^2]$.

Our goal is to assign a specific label to each $v$, drawn from $\mathcal{Y}$. We denote the indicator matrix $\bm{Z}\in\{0,1\}^{V\times Y}$, which means that $z_{vy}=1$ iff the label assigned to $v$ equals $y$. We want to infer $\bm{Z}$ given the visual feature matrix $\bm{X}$, the mapping $\psi$ and the temporal intervals $T_v, \ T_w$. 
\subsection{Clustering Model}

Our model is based on DIFFRAC \cite{NIPS2007_3269}, a discriminative clustering method. In particular, Bach and Harchaoui, in order to assign labels to unsupervised data, form a ridge regression loss function using a linear classifier ${f(\bm{x})=\bm{x}\bm{\omega}+\bm{b}}$, where $\bm{\omega} \in \mathbb{R}^{D \times Y}$ and $\bm{b} \in \mathbb{R}^{1 \times Y}$, which is optimized by the following:
\begin{equation}\label{objective}
	\min_{\bm{Z,\omega,b}}\frac{1}{2V}\|\bm{Z}-\bm{X\omega}-{\boldsymbol{1}_V}\bm{b}\|_F^2+\frac{\lambda}{2} Tr(\bm{\omega}^T\bm{\omega}),
\end{equation}
where $\lambda$ stands for the regularization parameter. Eq. \eqref{objective} can be solved analytically \wrt the classifier leading to a new objective function that needs to be minimized only \wrt the assignment matrix $\bm{Z}$: 
$\min_{\bm{Z}}\frac{1}{2}(\bm{ZZ}^T\bm{A}(\bm{X},\lambda))$, where $\bm{A}(\bm{X},\lambda)$ is a matrix that depends on the parameter $\lambda$ and the Gram matrix $\bm{XX^T}$, which can be replaced by any kernel (see \cite{NIPS2007_3269}). Relaxing the matrix $\bm{Z}\in\{0,1\}^{V\times Y}$ to $\bm{Z}\in[0,1]^{V\times Y}$, the objective becomes a convex quadratic function constrained by the following:
\begin{align}
	&\forall v \in \mathcal{V},\ \forall y \in \mathcal{Y},\ z_{vy}\geq 0 , \label{constraint2}\\
	&\bm{Z}{\cdot}{\boldsymbol{1}_Y} = {\boldsymbol{1}_V} , \label{constraint3}
\end{align}\\[-4mm]
where $z_{vy}$ denotes the $(v,y)$ element of matrix $\bm{Z}$.
\subsection{Weakly Supervised Methods}\label{weakly}
In order to incorporate in the model the weak supervision that the complementary modality provides, we have to resolve two kinds of ambiguities: \\
\noindent $\bullet$ Which visual object $v$ is described by each linguistic object $w$?\\
\noindent $\bullet$ Which label $y_w$ is implied from each $w$?
\subsubsection{Fuzzy Sets Multiple Instance Learning - FSMIL}
In an attempt to address the first question, similar to \cite{Bojanows13finding}, we assume that each $w$ should describe at least one of the $v$ that temporally overlaps with it. This leads to a multi-class MIL approach, where for each $w$ a bag-of-instances is created containing all the overlapping $v$: 
\begin{equation}\label{simple_sets}
\mathcal{V}_w=\{v \, | \, v\in\mathcal{V} \, , \, 
T_w\cap T_v\neq 0\}.
\end{equation}
We extend this framework in order to discriminate between visual objects with different temporal overlaps. In fact, the longer the overlap, the more likely it is for a visual object $v$ to be described by the corresponding linguistic $w$. For example, during a dialogue, in the video stream the camera usually focuses on the current speaker longer than the silent person, while the document mentions the first. Thus, we need to encode this observation on our MIL sets. This is done by defining a novel type of MIL sets using fuzzy logic (see Figure \ref{fig:fig3}). Each member of the set is accompanied by a value that demonstrates its membership grade : 
\begin{equation}\label{fuzzy_sets}
\mathcal{V}_w=\{(v,\mu_w(v)) \, | \, v\in\mathcal{V} \, , \, \mu_w(v)=g(\frac{|T_w\cap T_v|}{|T_v|})\},
\end{equation}
where $g$ is an increasing membership function with $g(0)=0, \ g(1)=1$. In addition, we note that, in order to compensate for the crude alignment mistakes, we can add a hyper-parameter $\epsilon$ that adjusts the linguistic object time interval as follows: ${\acute{T_w}=[t_w^1-\epsilon r, t_w^2 +\epsilon r]}$, where $r=|T_w|\mathbin{/}|\overline{T_w}|$ and $\overline{|T_w|}$ is the average value of $|T_w|$, over all $w$.

\subsubsection{Probabilistic Labels Multiple Instance Learning - PLMIL}\label{probabilistic}

As mentioned before, the labels extracted from the complementary modality involve a level of uncertainty. This happens due to the fact that the extraction procedure is a classification problem on its own. Solving this problem is equivalent to inferring the mapping $\psi$. Obtaining the label that the classifier predicts for each linguistic object $w$, renders the mapping deterministic, while obtaining the posterior probabilities that the classifier gives, renders it probabilistic. 

In this work, we use a probabilistic mapping using the posterior probabilities $\psi_w(y)=\mathbb{P}[y_w=y|w]$. In order to match them with the visual objects $v$, we perceive them as Probabilistic Labels (PLs). As mentioned in \cite{jin2003learning}, matching a PL to an instance that needs to be classified, accounts for an initial estimation of its true label. In our problem, we generalize the definition of \cite{jin2003learning}, matching PLs to bags-of-instances, meaning that at least one instance of the set should be described by this measure of initial confidence. In this case, the model's input data is formed as follows: $D=\bigcup_{w\in\mathcal{W}}(\mathcal{V}_w,\psi_w(\cdot))$.

We address the classification problem of text segments in an unsupervised manner. Specifically, we calculate the semantic similarity $s_{wy}$ of each $w$ with the linguistic representation of each label $y$ using the algorithm of \cite{han2013umbc}. We also apply a threshold $\theta$ to each similarity value in order to eliminate the noisy $w$ that do not imply any of the labels. Thus, for each $w$ we obtain a similarity vector $\bm{s}_w$, which is then normalized to constitute a p.m.f : ${\psi_w(y)=s_{wy}\mathbin{/}\sum_{\ell\in\mathcal{Y}}s_{w\ell}.}$ 
\subsubsection{Integration of the Weak Supervision in the Clustering Model}
In the MIL case each bag $\mathcal{V}_w$ is matched to a single label $y$ and is represented by the following constraint: 
\vspace{-0.1cm}
\begin{equation}
\forall w \in \mathcal{W} \ ,\ \sum_{v \in \mathcal{V}_w}z_{vy}\geq 1.
\end{equation} \\ [-2mm]
For the purpose of accounting for noise, slack variables are used to reformulate both the objective function and the constraints:
\vspace{-0.1cm}
\begin{align}\label{slacks}
   &\min_{\bm{Z},\bm{\xi}}Tr(\bm{ZZ}^T\bm{A}(\bm{X},\lambda)) + \kappa\sum_{w \in \mathcal{W}} \xi_{w}^2,\\
   &\textrm{s.t} \ \ \ \forall w \in \mathcal{W} \ ,\ \sum_{v \in \mathcal{V}_w}z_{vy}\geq 1 - \xi_{w}.
\end{align}\\ [-2mm]
In our FSMIL, we intend to add different weights to the elements of each bag depending on the membership grade:
\vspace{-0.1cm}
\begin{align}\label{fuzzy_uncertainty_constraint}
	&\forall w \in \mathcal{W} \ ,\ \sum_{v \in \mathcal{V}_w}z_{vy}\mu_w(v)\geq 1-\xi_w.
\end{align} \\ [-2mm]
For the PLMIL case, let $\mathcal{Y}_w\subseteq\mathcal{Y}$ be the set of the labels for which the p.m.f $\psi_w(y)$ is non-zero. For each label  $y \in \mathcal{Y}_w$ we construct a constraint formed as in \eqref{fuzzy_uncertainty_constraint}, \ie :
\vspace{-0.1cm}
\begin{align}\label{probabilistic_label_constraint}
	&\forall w \in \mathcal{W} \ , \ \forall y \in \mathcal{Y}_w \ , \ \sum_{v \in \mathcal{V}_w}z_{vy}\mu_{w}(v)\geq 1-\xi_{wy}.
\end{align} \\ [-2mm]
The discrimination between the various labels of $\mathcal{Y}_w$ is carried out by the slack variables. In particular, we rewrite the objective function as follows:
\vspace{-0.1cm}
\begin{align}\label{weightedslacks}
   &\min_{\bm{Z},\bm{\xi}}Tr(\bm{ZZ}^T\bm{A}(\bm{X},\lambda)) + \kappa\sum_{w \in \mathcal{W}} \sum_{y \in \mathcal{Y}_w} f(\psi_w(y))\xi_{wy}^2,
\end{align} \\ [-2mm]
where $f(\cdot)$ is an increasing function that transforms probabilities to slack variable weights. In this way, we manage to relax the constraints inversely proportional to the probability of the corresponding label. As a result, a constraint is harder to be violated as long as the probability is high.

\noindent\textbf{Rounding:} Similarly to \cite{Bojanows13finding} we choose a simple rounding procedure for $\bm{Z}$ that accounts for taking the maximum values along its rows and replacing it with 1. The rest of the values are replaced with 0.

\section{Experiments}\label{section:exp}

\subsection{Dataset}
The COGNIMUSE database \cite{data} is a video-oriented database multimodally annotated with audio-visual events, saliency, cross-media relations and emotion \cite{CognimuseDB}. It is a generic database that can be used for  event detection and summarization, as well as audio-visual concept recognition. Other existing databases such as the MPII-MD \cite{rohrbach2015dataset}, the M-VAD \cite{torabi2015using}, the MSVD \cite{chen2011collecting}, the MSR-VTT \cite{xu2016msr}, the VTW \cite{zeng2016generation}, the TACoS \cite{regneri2013grounding,rohrbach2013translating}, the TACoS Multi-Level \cite{rohrbach2014coherent} and the YouCook \cite{DaXuDoCVPR2013} are not annotated in terms of specific visual concepts, but in terms of sentence descriptions. Moreover, the datasets used in \cite{cour2009learning,conf/bmvc/EveringhamSZ06,ramanathan2014linking,conf/cvpr/SivicEZ09} are only annotated with human faces. Finally, the Hollywood2 \cite{marszalek2009actions} and the Casablanca \cite{Bojanows13finding} datasets were not sufficient for the action recognition task, due to the fact that only automatically collected labels from the text are provided rather than the text itself. Regarding Casablanca, it was not possible to apply our FMSIL method for the face recognition task either, as we observed that, in this movie, each face track and each sentence have either zero overlap ($T_v\cap T_w=\emptyset$) or the face track's temporal interval is a subset of that of the sentence (hence $T_v\cap T_w= T_v$), which  always yields a zero or unitary membership grade for the FSMIL case. On the contrary, the COGNIMUSE database consists of long videos that are continuously annotated with action labels and are accompanied by texts in a raw format. In addition, we manually annotated the detected face tracks in order to evaluate the face recognition task. All the above, render COGNIMUSE more relevant and useful for the tasks that we are dealing with. In this work, we used 5 out of the 7 annotated 30-minutes movie clips, which are: “A Beautiful Mind” (BMI), “Crash” (CRA), “The Departed” (DEP), “Gladiator” (GLA) and “Lord of the Rings - the Return of the King” (LOR).

\begin{table*}[t]
\begin{center}
\resizebox{0.70\linewidth}{!}{
\begin{tabular}{|l|c|c|c||c|c|c|c||c|}
\hline
\textbf{Set}&\multicolumn{3}{c||}{Development}&\multicolumn{4}{c||}{Test}&\multicolumn{1}{c}{All}\vline\\
\hline\hline
&\textbf{DEP}&\textbf{LOR}&\textbf{MAP}&\textbf{BMI}&\textbf{CRA}&\textbf{GLA}&\textbf{MAP}&\textbf{MAP}\\\hline
\textbf{Text+MIL}&0.433&0.656&0.544&0.551&0.434&0.437&0.474&0.502\\\hline
\textbf{SIFT+MIL} \cite{Bojanows13finding}&0.630&0.879&0.755&0.724&0.644&0.681&0.683&0.711\\\hline
\textbf{SIFT+FSMIL }&0.693&0.881&0.787&0.770&0.691&0.746&0.736&0.756\\\hline
\textbf{VGG+MIL}&0.834&\textbf{0.954}&0.894&0.825&0.696&0.830&0.784&0.828\\\hline
\textbf{VGG+FSMIL (Ours)}&\textbf{0.864}&0.952&\textbf{0.908}&\textbf{0.857}&\textbf{0.731}&\textbf{0.901}&\textbf{0.830}&\textbf{0.861}\\\hline
\hline
\textbf{\cite{miech2017learning}+VGG: fg}&0.788&0.898&0.843&0.666&0.479&0.577&0.574&0.682\\\hline
\textbf{\cite{miech2017learning}+VGG+FSMIL: fg}&\textbf{0.810}&\textbf{0.913}&\textbf{0.862}&\textbf{0.696}&\textbf{0.505}&\textbf{0.651}&\textbf{0.617}&\textbf{0.715}\\\hline\hline
\textbf{\cite{miech2017learning}+VGG: bg}&\textbf{0.185}&\textbf{0.189}&\textbf{0.187}&\textbf{0.304}&0.047&\textbf{0.052}&0.134&0.155\\\hline
\textbf{\cite{miech2017learning}+VGG+FSMIL: bg}&0.184&\textbf{0.189}&\textbf{0.187}&0.269&\textbf{0.278}&0.038&\textbf{0.195}&\textbf{0.192}\\\hline
\end{tabular}}
\end{center}
\vspace{-0.4cm}
\caption{The Average Precision (AP) scores of each movie in the Development and the Test Set for the Face Recognition Task. The Mean Average Precision (MAP) calculated in the two sets separately and for the database as a whole is also shown.}
\vspace{-0.2cm}
\label{face_table}
\end{table*}

\subsection{Implementation}
\noindent\textbf{Detection and Feature Extraction}: We spatio-temporally detect and track faces similarly to \cite{Bojanows13finding}, where face tracks are represented by SIFT descriptors and the kernels are computed separately for each facial feature taking into account whether a face is frontal or profile. Contrary to this, we use deep features extracted by the last fully connected layer of the VGG-face pre-trained CNN \cite{Parkhi15}, while a single kernel is computed on each pair of face tracks regardless to the faces' poses. Similarly to \cite{Bojanows13finding,conf/bmvc/EveringhamSZ06,conf/cvpr/SivicEZ09} the kernel applied is a min-min RBF. For the problem of action recognition, we use the temporal boundaries provided by the dataset. We represent them through the C3D pre-trained CNN, following the methodology stated in \cite{DBLP:journals/corr/TranBFTP14} .

\noindent\textbf{Label Mining from Text}: Prior to applying the label extraction algorithms, we perform a crude alignment between the script and the subtitles through a widely used DTW algorithm \cite{conf/bmvc/EveringhamSZ06}. The label set for the face recognition task is defined using the cast list of each movie (this information was downloaded from the Website TMDB \cite{tmdb}). The character labels are then extracted using regular expression matching, where the query expressions are the names included in the cast list. We define the label set for the action recognition task using a subset of the total classes of the COGNIMUSE database. We locate the linguistic objects $w$ by composing short sentences constituted by each sentence's verb as well as words that are linked to the verb through specific dependencies, such as the direct object and adverbs. We use the toolbox CoreNLP \cite{manning-EtAl:2014:P14-5} in order to perform the document's dependency parsing. Finally, we calculate the semantic similarities on every label - short sentence pair applying an off-the-shelf sentence similarity algorithm \cite{han2013umbc}. This comprises a hybrid approach between Latent Semantic Analysis (LSA) and knowledge from WordNet \cite{miller1995wordnet}. The similarities that do not exceed a specific threshold $\theta$, experimentally set to 0.4, are discarded. 

\subsection{Learning Experiments}

In the following experiments we evaluate our methods on the tasks of (i) face and (ii) action recognition. For the FSMIL setting, after extensive experimentation with a variety of $\Gamma$ and $S$-shaped membership functions (\eg sigmoid, linear, piecewise quadratic, cubic), we selected a specific $\Gamma$-shaped function, using as criteria the performance and sensitivity to hyperparameters, that is defined as follows:
\vspace{-0.2cm}
\begin{equation}\label{mapping_function}
g(x)= \begin{cases} 
      0 & x\leq a \\
      \frac{k(x-a)^2}{1+k(x-a)^2} & a\leq x\leq 1 \\
   \end{cases}
\end{equation}\\[-3mm]
where $a$ is the membership threshold and $k$ is a parameter that controls how abrupt the increase above the threshold will be. We need to assign $k$ a large value (above 1000) in order to have $g(1)=1$. For those values there are no significant changes in the results. We tune the hyperparameters $a$ and $\epsilon$ on the Development set independently for the two tasks, yielding $a = 0.2$, $\epsilon = 0$ for task (i) and $a = 0.1$, $\epsilon = 100$ for task (ii). For the PLMIL setting, we observed that, in the COGNIMUSE dataset, the best results were obtained when weights were close to 1 (candidate labels), thus the mapping function $f$ is again given by \eqref{mapping_function}, setting $a$ to 0 and $k$ to a large value as previously stated. The hyperparameter sensitivity was assessed on the Test set and we saw that the deviations from the reported results were small, thus the sensitivity is assumed to be low. Moreover, the chosen values demonstrate almost optimal performance on the Test set as well, thus there was no overfitting on the Development set. The source code of the entire system and experiments (including results on hyperparameter sensitivity), as well as precomputed features can be found at \cite{code}.

\vspace{-0.3cm}
\subsubsection{Face Recognition}\label{face_recognition}
We evaluate each method's performance using the Average Precision (AP) previously used in \cite{Bojanows13finding,cour2009learning}. We compare our model (VGG+FSMIL) to the methodology of Bojanowski \etal \cite{Bojanows13finding} - that has outperformed other weakly supervised methods, such as \cite{cour2009learning} and \cite{conf/cvpr/SivicEZ09} - as well as with other baselines described next: \\ [-2mm]

\noindent $1$. Text+MIL: We solve the problem by minimizing only the factor related to the slack variables. This method, converges to an optimum that satisfies the constraints posed by the text, without taking into account the visual features. The constraints are formed using the simple MIL setting. \\
\noindent $2$. SIFT+MIL \cite{Bojanows13finding}: The algorithm of Bojanowski \etal that uses SIFT descriptors as feature vectors and simple MIL setting, without taking into account the temporal overlaps, namely the bags are constructed as noted by \eqref{simple_sets}. \\
\noindent $3$. SIFT+FSMIL: Our proposed learning method implemented with SIFT descriptors. \\
\noindent $4$. VGG+MIL: The algorithm of Bojanowski \etal implemented with VGG-face descriptors. \\
\noindent $5$. \textbf{VGG+FSMIL (Ours)}: The proposed learning method implemented with VGG-face descriptors.\\ [-2mm]
 
The comparative results for each movie are shown in Table \ref{face_table}. As it can be clearly seen, our method demonstrates superior performance than \cite{Bojanows13finding} concerning every case.

First of all, the inferior performance of the Text+MIL method shows the inefficiency of using only textual information in tackling the problem. The higher accuracy accomplished by the methods implemented with VGG proves the benefits of deep learning over hand-crafted features as means of representing faces. Moreover, incorporating the information given by the overlaps of visual and linguistic objects, we improve the accuracy regardless of the nature of the representation. In particular, due to the fact that our method reduces the ambiguity in each bag-of-instances, we outperform the baseline even without the use of deep features. As expected, the combination of the above (VGG and FSMIL) shows the highest accuracy. This can be easily explained as each one of the methods improves different aspects of the learning procedure. 

In the comparisons above, we discard background (bg) characters and optimize on foreground (fg). Additionally, in Table \ref{face_table}, we compare our method to \cite{miech2017learning}, that has achieved state-of-the-art results by using a bg class constraint. In particular, tracks that have no assigned weak label are collected in a bag and a percentage of them, determined by a hyperparameter $\alpha_2$ (we set it to 0.6), is enforced by a constraint to be classified as bg. We observe that when we incorporate FSMIL, our method always outperforms \cite{miech2017learning} on fg characters and provides either similar, or better results on bg.

\vspace{-0.3cm}
\subsubsection{Action Recognition}
\begin{figure*}[t]
\centering
    \begin{subfigure}[h]{0.26\textwidth}
    \centering
        \includegraphics[width=\textwidth]{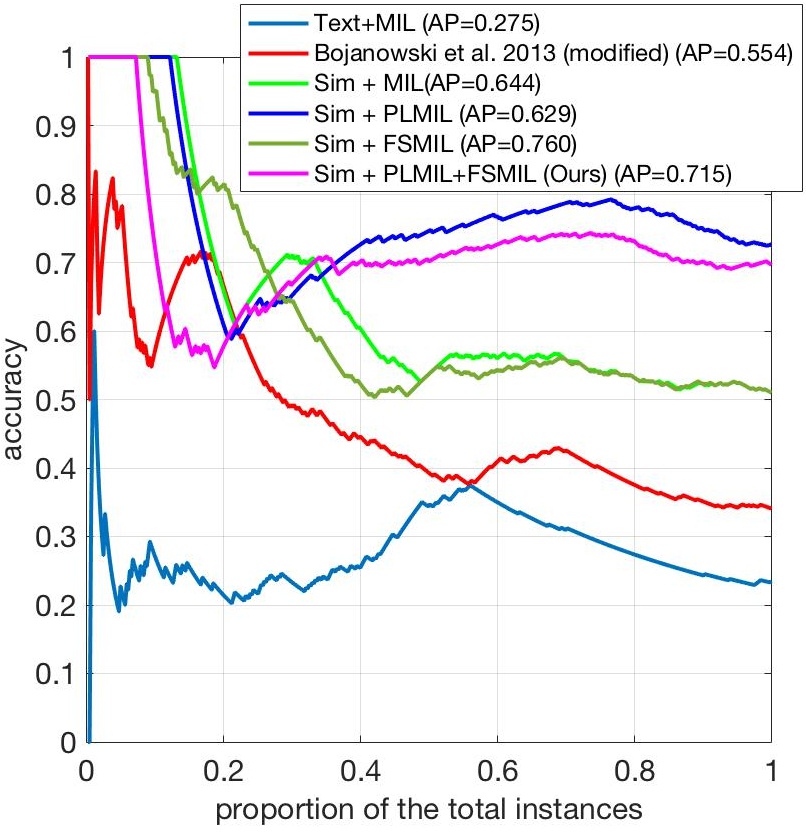}
                 \caption{}
    \end{subfigure}
    \begin{subfigure}[h]{0.26\textwidth}
    \centering
        \includegraphics[width=\textwidth]{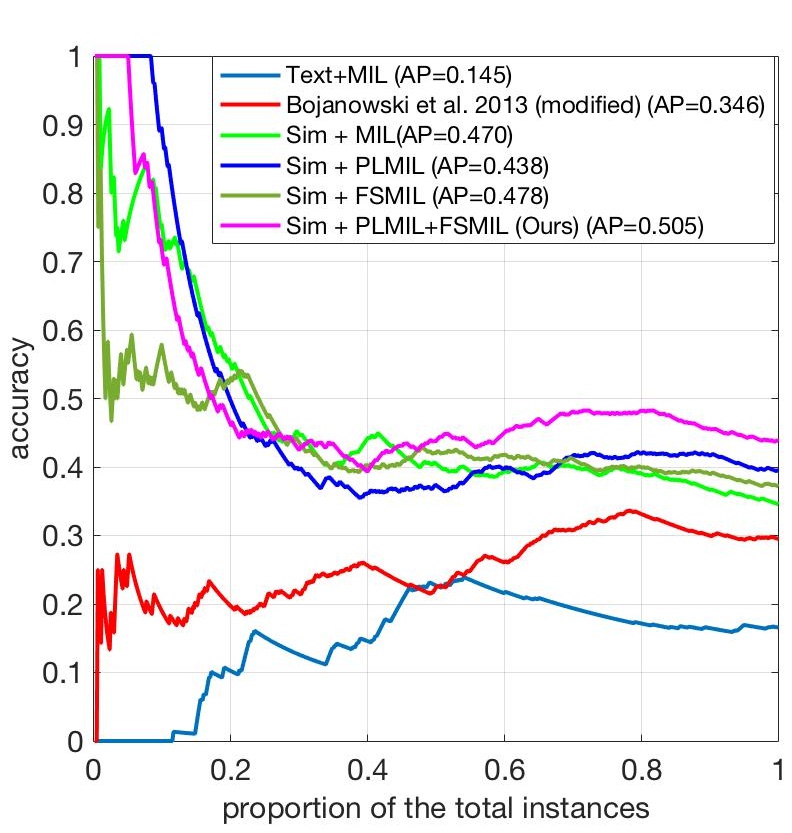}
                 \caption{}
    \end{subfigure}
    \begin{subfigure}[h]{0.26\textwidth}
    \centering
        \includegraphics[width=\textwidth]{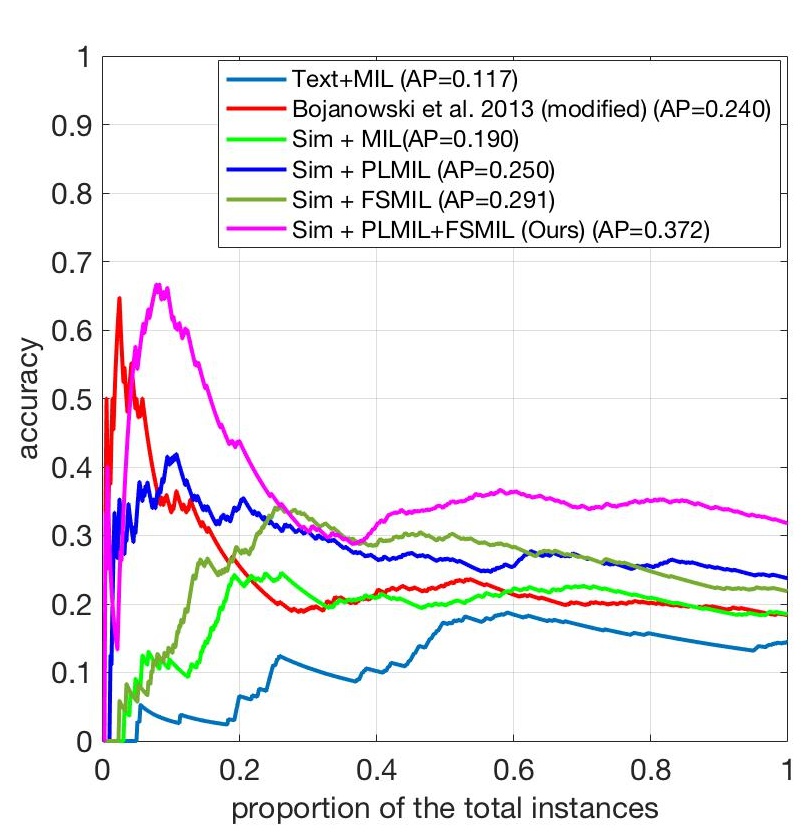}
             \caption{}
           \end{subfigure} \\ \vspace{-0.1cm}
    \begin{subfigure}[h]{0.26\textwidth}
        \centering
        \includegraphics[width=\textwidth]{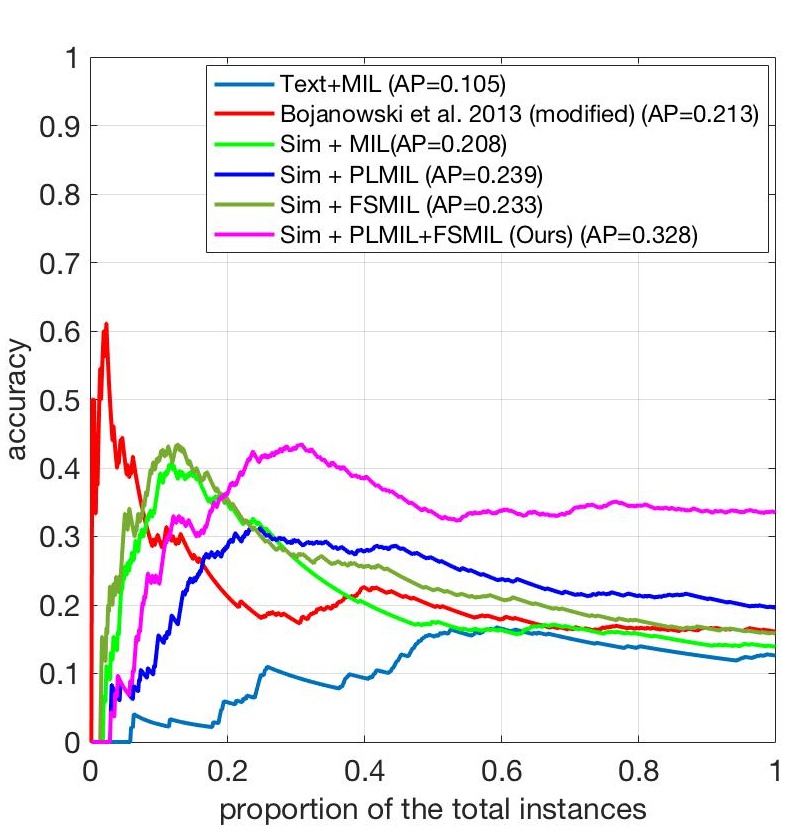}
         \caption{}
    \end{subfigure}
    \begin{subfigure}[h]{0.26\textwidth}
        \centering
        \includegraphics[width=\textwidth]{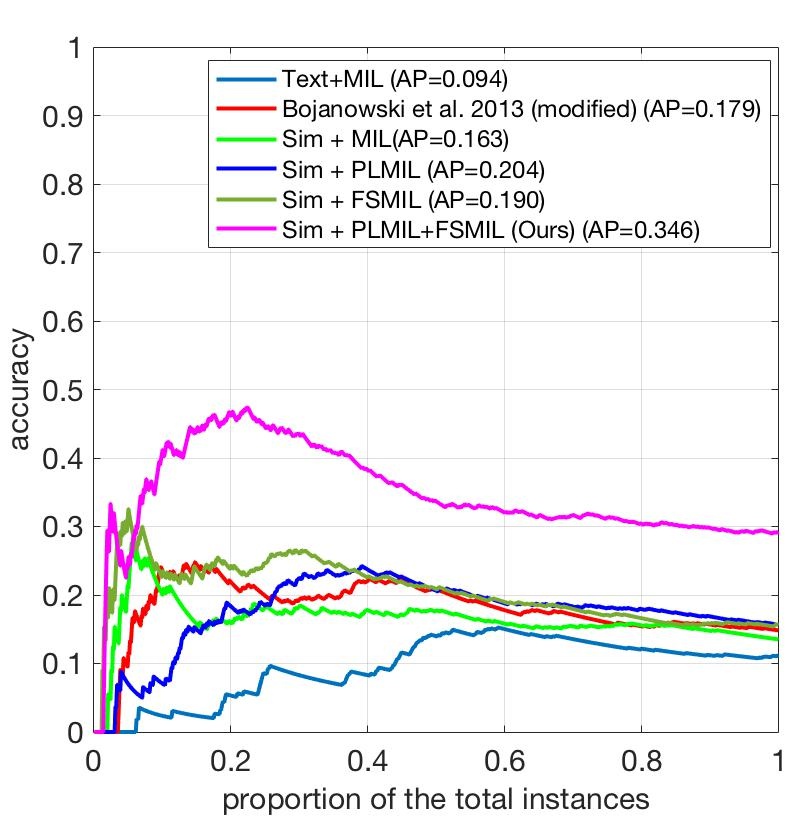}
        \caption{}
    \end{subfigure}
    \vspace{-0.3cm}
    \caption{The curves show the per sample accuracy plotted against the proportion of total instances, concerning the whole dataset. Each figure corresponds to a different experiment concerning the number of classes. Note that as the number of classes increases, the model that combines our methods greatly outperforms each one of them individually as well as the baseline. Please see color version for better visibility.}
    \vspace{-0.3cm}
    \label{fig:precision_recall}
\end{figure*}

Regarding the task of action recognition, several experiments were carried out for each movie, while changing the cardinality of the label set. In particular, the performance is evaluated using the 2, 4, 6, 8 and 10 most frequent action classes. The evaluation is performed using the Mean AP metric, which stands for averaging the APs over each movie set. The results are demonstrated both for the Development and the Test set in Table \ref{table_action}. We also illustrate the performances of the methods on the whole dataset with the \textit{per sample accuracy} vs \textit{proportion of total instances} curves of Figure \ref{fig:precision_recall}.

We choose as baseline the aforementioned Text+MIL, as well as a similar methodology to Bojanowski \etal \cite{Bojanows13finding}. In this experiment, we focus on the different ways of learning from the text, rather than the visual features, thus in all cases we use the C3D descriptor for the representation of actions. The methods compared are: \\ [-2mm]

\noindent $1$. Text+MIL: Same as the one described in section \ref{face_recognition}. The action labels are extracted by locating the sentences that are \textbf{semantically identical} to one of the labels of the set $\mathcal{Y}$ (similarity  = 1). \\
\noindent $2$. MIL (\cite{Bojanows13finding} modified): The learning algorithm of Bojanowski \etal mentioned in \ref{face_recognition}. We replace the dense trajectories descriptors with C3D. Again, we use only the sentences that are \textbf{semantically identical} to some label. \\ 
\noindent $3$. Sim+MIL: The same learning algorithm, but labels are extracted from sentences that are \textbf{semantically similar} to one of the labels of the set $\mathcal{Y}$ ($\theta\leq\textrm{similarity}\leq1$). Each sentence is assigned a single label, the one with the maximum similarity. \\
\noindent $4$. Sim+PLMIL: Our PLMIL method. We assign a probabilistic label to each sentence. \\
\noindent $5$. Sim+FSMIL: Our FSMIL method. We construct the bags-of-instances as fuzzy sets. \\
\noindent $6$. \textbf{Sim+FSMIL+PLMIL (Ours)}: The combination of our contributions using semantically similar sentences, probabilistic labels and fuzzy bags-of-instances. \\ [-2mm]

\begin{table}[t]
\begin{center}
\resizebox{\linewidth}{!}{
\begin{adjustbox}{max width=0.50\textwidth}
\begin{tabular}{|l|c|c|c|c|c|}
\hline
\textbf{Number of Classes} & \textbf{2}&\textbf{4}&\textbf{6}&\textbf{8}&\textbf{10}\\\hline
\textbf{Set} & \multicolumn{5}{|c|}{\textbf{Development}} \\ \hline
\textbf{Text+MIL}&0.566&0.315&0.253&0.083&0.089\\\hline
\textbf{MIL(modified) \cite{Bojanows13finding}}&0.638&0.420&0.283&0.266&0.194\\\hline
\textbf{Sim+MIL}&0.837&0.299&0.243&0.339&0.202\\\hline
\textbf{Sim+PLMIL}&0.837&0.304&0.308&0.348&0.229\\\hline
\textbf{Sim+FSMIL}&\textbf{0.945}&0.614&0.435&0.383&0.317\\\hline
\textbf{Sim+FSMIL+PLMIL (Ours)}&\textbf{0.945}&\textbf{0.617}&\textbf{0.520}&\textbf{0.491}&\textbf{0.450}\\\hline \hline
\textbf{Set} & \multicolumn{5}{|c|}{\textbf{Test Set}} \\ \hline
\textbf{Text+MIL}&0.306&0.199&0.188&0.134&0.147\\\hline
\textbf{MIL(modified) \cite{Bojanows13finding}}&0.405&0.180&0.184&0.189&0.212\\\hline
\textbf{Sim+MIL}&0.631&\textbf{0.591}&0.182&0.094&0.129\\\hline
\textbf{Sim+PLMIL}&0.585&0.564&0.298&0.140&0.148\\\hline
\textbf{Sim+FSMIL}&\textbf{0.792}&0.458&0.232&0.168&0.146\\\hline
\textbf{Sim+FSMIL+PLMIL (Ours)}&0.692&0.521&\textbf{0.299}&\textbf{0.249}&\textbf{0.270} \\ \hline
\end{tabular}
\end{adjustbox}
}
\end{center}
\vspace{-0.4cm}
\caption{The Mean Average Precision (AP) scores over the \textbf{Development} and \textbf{Test} set for five independent experiments for the Action Recognition Task.}
\vspace{-0.4cm}
\label{table_action}
\end{table}

First note that the proposed combined model demonstrates superior performance over the Text + MIL baseline, confirming the importance of using visual information, as previously mentioned in \ref{face_recognition}. Higher performance is also reported over the baseline of \cite{Bojanows13finding} in every case, leading to an improvement of 20\% -- 30\% in the Development set and 6\% -- 34\% in the Test set. Moreover, Figure \ref{fig:precision_recall} shows that it outperforms all methods in the whole dataset, except for the case of two classes. Next, we examine each of our contributions independently.

The method of extracting labels through similarity measurements outperforms the baseline mainly when the number of classes is small (2-4), as shown in Table \ref{table_action}. In this case, the concepts implied by the labels, in terms of semantics, are rarely confused, hence most of the similarity measurements produce correct labels.  However, as this number increases the Sim+MIL method does not prove very efficient on its own. A possible explanation is that the semantically identical labels of the baseline usually consist of a more clean set, while the confusion introduced to the model with semantically similar labels rises. As a result, despite the fact that a small amount of bags-of-instances are annotated, the baseline algorithm will still be able to make a few correct predictions with large confidence. This is illustrated in Figure \ref{fig:precision_recall} (c) and (d), where the most confident predictions of the baseline are accurate, contrary to those of Sim+MIL.

This confusion is compensated partially by either PLMIL or FSMIL. Regarding the first one, when the classes are few, a sentence is rarely similar to more than one concepts, hence the labels are mainly deterministic. However, modeling labels in a probabilistic way achieves better disambiguation of the sentences' meanings as the number of classes grows larger, which is proved by the fact that Sim+PLMIL outperforms Sim+MIL for 6-10 classes in both sets. As far as the FSMIL is concerned, this method is expected to perform better on its own for the reasons mentioned in section \ref{face_recognition}, regardless of the number of classes. Indeed, Sim+FSMIL outperforms Sim+MIL in most of the cases.

Interestingly, the combination of our contributions manages to outperform the baseline, even if none of them could do so independently. This can be explained by the fact that the algorithm leverages each one of them to resolve different kinds of ambiguities. Regarding the lower results in the Test set compared to the Development, we noticed that the scripts of the test movies are not sufficiently aligned to the videos, while a significant amount of actions occur in the background, consequently are not described in the text.

\section{Conclusion}
In this work we tackled the problem of automatically learning visual concepts by combining visual and textual information. We proposed two novel weakly supervised techniques that can be easily generalized to other Multimodal Learning tasks, that efficiently deal with temporal ambiguities (FSMIL), as well as semantic ones (PLMIL). Contrary to previous work, we acquire richer information from the text using semantic similarity. We evaluated our models on the COGNIMUSE dataset, containing densely annotated movies accompanied by their scripts. Our techniques provide significant improvement over a state-of-the-art weakly supervised method, in both face and action recognition tasks. Regarding our future work, we plan to extend our uni-directional model to a bi-directional, where information will flow from text to video and vice-versa, jointly learning visual and linguistic concepts. Finally, the generality of our formulation motivates us in exploring its potential in learning from other modalities such as the audio channel. 

\vspace{0.2cm}
\noindent \textbf{Acknowledgements:} This work was partially supported by EU projects Babyrobot/687831 and i-Support/643666.
\newpage
{\small
\bibliographystyle{ieee}
\bibliography{egbib}
}

\end{document}